\documentclass[final]{cvpr}

\usepackage{times}
\usepackage{epsfig}
\usepackage{graphicx}
\usepackage{amsmath}
\usepackage{amssymb}
\usepackage[toc,page]{appendix}
\usepackage{booktabs}
\usepackage{float}
\RequirePackage{ifpdf}
\usepackage[pagebackref=true,breaklinks=true,colorlinks,bookmarks=false]{hyperref}

\usepackage{xcolor}
\def\confYear{CVPR 2021}

\begin{document}
\ifpdf
\else
\begin{center}
\textbf{Warning:} The DVI version of this paper may be corrupted.  If possible, use the PDF version.
\end{center}
\fi

%%%%%%%%% TITLE
\title{MakeupBag: Disentangling Makeup Extraction and Application}
\author{Dokhyam Hoshen\\
Lightricks\\
\\

}

\maketitle

%%%%%%%%% ABSTRACT
\begin{abstract}
This paper introduces MakeupBag, a novel method for automatic makeup style transfer. Our proposed technique can transfer a new makeup style from a reference face image to another previously unseen facial photograph. We solve makeup disentanglement and facial makeup application as separable objectives, in contrast to other current deep methods that entangle the two tasks. MakeupBag presents a significant advantage for our approach as it allows customization and pixel specific modification of the extracted makeup style, which is not possible using current methods. Extensive experiments, both qualitative and numerical, are conducted demonstrating the high quality and accuracy of the images produced by our method. Furthermore, in contrast to most other current methods, MakeupBag tackles both classical and extreme and costume makeup transfer. In a comparative analysis, MakeupBag is shown to outperform current state-of-the-art approaches. 
\end{abstract}

%%%%%%%%% BODY TEXT
\section{Introduction}

\begin{figure}[!ht]

\begin{tabular}{ccc}
\includegraphics[width = 0.14\textwidth,height=0.14\textwidth]{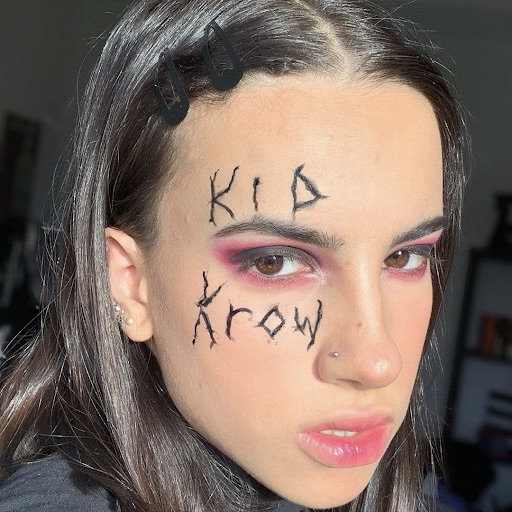}&
\includegraphics[width = 0.14\textwidth,height=0.14\textwidth]{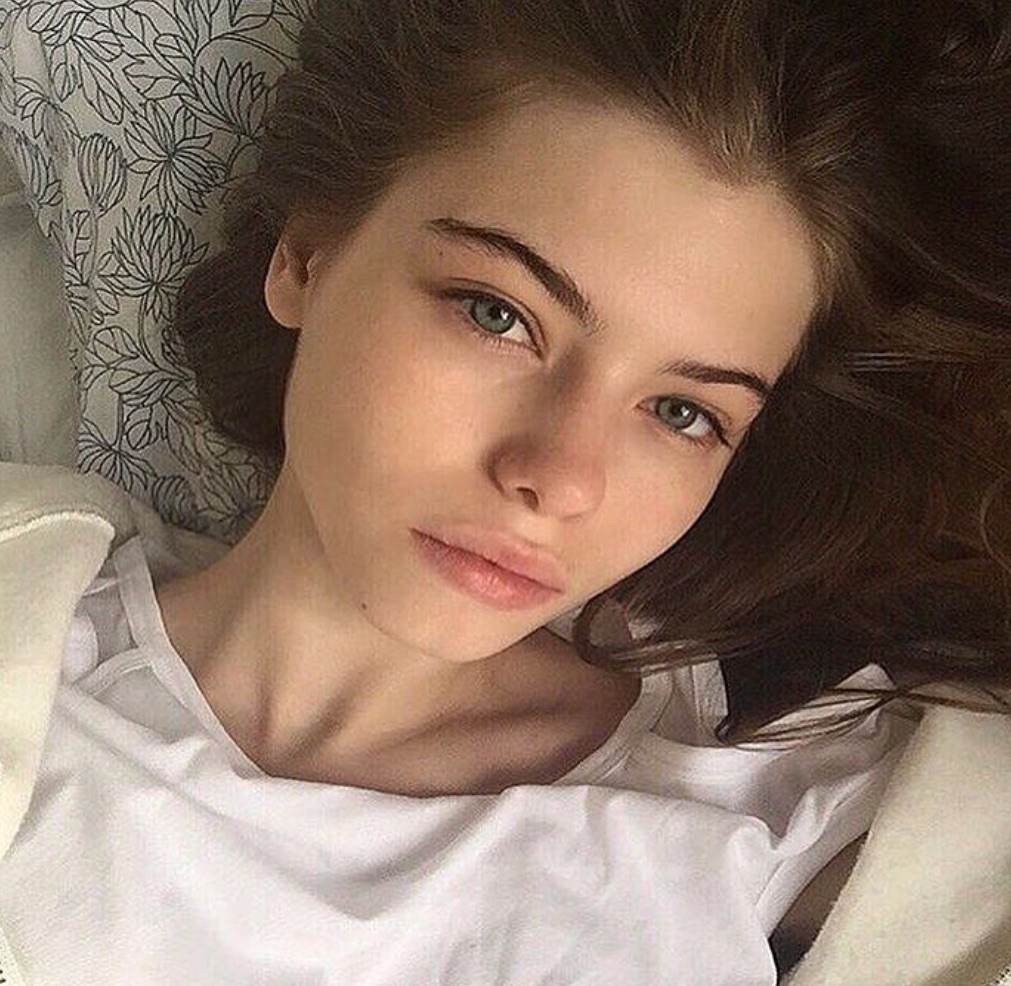}&

\includegraphics[width = 0.14\textwidth,height=0.14\textwidth]{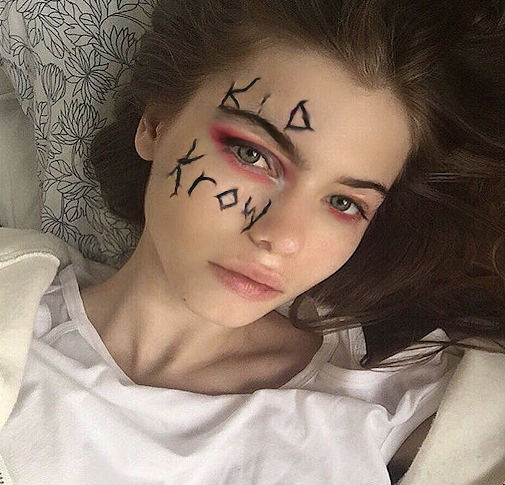} \\
Reference & Target & Output: \\ & & MakeupBag

\end{tabular}
\caption{ Example of MakeupBag's typical inputs and output. Presented are a triplet of (reference, target, output) that makeup transfer aims to produce. The reference image depicts a specific makeup style that we would like to transfer to the target image, which is typically a photo of a face with no makeup. As can be appreciated, MakeupBag provides the desired output: the target image is unmistakably wearing the makeup style seen in the reference image. }
\label{fig:front_page_fig}
\end{figure}

We introduce MakeupBag, a method of transferring a makeup style depicted in a reference photo to an arbitrary target face (See Figure \ref{fig:front_page_fig}). We approach this task as consisting of two steps: i) Makeup extraction: disentangling the makeup style observed in a facial photograph from the wearer's identity. ii) Makeup application: applying a particular makeup style provided as an input to a target facial photograph. 

The usefulness of automatic makeup transfer has attracted much research interest from academia and industry. Current virtual makeup application systems either permit the user to choose a style from a given set or enable the user to create their own makeup style digitally. Both options are disadvantageous as they provide low quality results, are time consuming or have limited style options.

 The task of automatic makeup transfer is however very challenging, as the makeup signal is noisy and is affected by different lighting, skin textures and colors. The task has initially been tackled by non-data driven methods, but the quality of results was insufficient for industrial applications. Recently, data driven methods have been employed for this task. A typical supervised learning method would require supervised aligned facial triplets, including two photographs of a person with and without a particular makeup style and a photograph of another person wearing the same makeup style. Unfortunately, current available datasets have at best unaligned before and after makeup pictures and more commonly, datasets can be produced that consist of unpaired images (different identities) of people with and without makeup. The task therefore requires unsupervised learning solutions. The difficulty of the task is compounded by the variety of makeup styles (mild through extreme), the details of the styles and their correlation to external factors such as the identity of the wearer and lighting. 

 Previously, most data driven methods have addressed this task by using unpaired domain transfer methods that solve the tasks of makeup removal and  makeup application jointly. This is disadvantageous for the following reasons: 
 \begin{itemize}
     \item The user has less direct control over the makeup applied beyond specification of the reference image. 
     \item Current unpaired methods can only operate at low image resolution.
     \item In the cases where the model operates on patches of the face, areas around the patch edges can have displeasing artifacts.
     \item Manipulation of the makeup style is very limited, enabling at most interpolation in latent space and not actual pixel-level editing.
 \end{itemize}

 In this paper we present MakeupBag. We first present an algorithm for extracting the makeup style from a facial image (the extraction phase). MakeupBag tackles makeup extraction as a makeup segmentation task, thus the output of our extraction method is an alpha segmentation mask. To our knowledge, supervised datasets for makeup segmentation are not available, making the Makeup segmentation phase more challenging. We then present a  GAN-based approach for applying a makeup style on an arbitrary face (the application phase). Our proposed method has the following advantages:
 \begin{itemize}
     \item As MakeupBag is not based on unpaired image translation methods, it can operate in higher resolutions and on the entire facial area.
     \item Since MakeupBag does not rely on division of the face into patches, it produces high quality results on makeup styles that appear on multiple facial regions, as is common in extreme makeup styles.
     \item MakeupBag enables editing of the makeup style so that only the desired areas are transferred. 
 \end{itemize}

%-------------------------------------------------------------------------
\section{Previous Work}

\subsection{Image Generation} 

Generating realistic images has long been the goal of many applications such as: domain transfer problems (Photos-to-emojis \cite{taigman2016unsupervised}, bags-to-shoes \cite{isola2017image} , zebras-to-horses \cite{zhu2017unpaired} ), generating synthetic datasets \cite{radford2015unsupervised}, text-to-images \cite{reed2016generative} \cite{zhang2017stackgan}, pose and expression manipulation \cite{ma2017pose}, photo editing (Face aging \cite{antipov2017face}, manipulating hair styles \cite{perarnau2016invertible}, accessories and facial attributes manipulation \cite{brock2016neural} \cite{liu2016coupled}), super-resolution \cite{ledig2017photo}, photo inpainting \cite{pathak2016context}, to list a few. This field has been revolutionized by Generative Adversarial Networks (GANs)\cite{goodfellow2014generative}, a method introduced by Goodfellow \textit{et al}. GANs have provided credible and aesthetic results to many of the applications listed above. In Pix2PixHD \cite{wangpix2pixhd}, Wang \textit{et al.} showed that they can produce high resolution (up to 2048x1024 resolution) images on paired data. As one of the main challenges for makeup transfer is producing realistic makeup images, this approach can benefit the makeup transfer application although the main obstacle in the field is the lack of supervised data. 

\subsection{Unsupervised Domain Transfer and Disentanglement}
In applications such as ours where supervision is scarce, unpaired GAN  methods are frequently applied. The seminal work is CycleGAN \cite{zhu2017unpaired} where Zhu \textit{et al.} introduced a reconstruction loss for unpaired domain transfer, thus aiming to eliminate the need for supervision in domain transfer problems. However, the reconstruction loss has proven to be unstable and very tricky to train. Furthermore, it is limited by image size and is unable to produce high quality, full size results.

In this setting of unsupervised or semi-unsupervised learning, much research has concentrated on disentanglement. Certain methods have disentangled different features (pose, expression) with various approaches such as Liu \textit{et al. } \cite{liu2017unsupervised}, Huang \textit{et al. } \cite{huang2018multimodal}, Fu  \cite{fu2019high}. These methods aim to learn optimal encoding for the latent variable that they try to disentangle. MUNIT \cite{huang2018multimodal} has shown success in domain transfer (e.g., cats to dogs) and therefore has been used for similar applications as cycleGAN \cite{zhu2017unpaired}. These applications are relevant to our problem as they too tackle the problem of transferring an entangled attribute from one image to the other. They differ in the type of attributes that they transfer, normally dealing with global attributes rather than local ones as in makeup transfer.  

\subsection{Makeup Transfer Methods}
Makeup transfer incorporates two problems: extraction of the makeup from a reference facial image and application of the makeup to a target no-makeup face. The makeup must possess a similar style as the original reference image while blending naturally with the no makeup target image, providing the face depicted in the target image with a makeup style resembling the reference image. Previous work on this problem has approached it as an unsupervised domain transfer problem, where makeup transfer is represented as style transfer. The objective is to transfer an image from  "without makeup" domain to "with makeup" domain. The difference is that in order to make this transformation, the network must have a reference image to restrict the makeup to a specific style in the makeup domain. In fact, the inverse transformation (as in makeup removal) has shown good results with a classic CycleGAN \cite{zhu2017unpaired}.  

The first to propose a GAN-based method for makeup transfer was Li \etal with BeautyGAN \cite{li2018beautygan}. The model expected multiple inputs (target and reference) and had a makeup loss that performs histogram matching for different facial segments. BeautyGlow \cite{gu2019ladn} proposed a similar framework in which Gu \etal used GLOW feature encoding to attempt to disentangle makeup from the identity of the wearer. Using the encoding, BeautyGlow showed they can  control the level of applied makeup.

Chang \textit{et al.} introducing pairedCycleGAN \cite{chang2018pairedcyclegan} used a cyclic loss as in CycleGAN, restricting the model to a specific makeup style. This method added an asymmetric loss where they warped the makeup image to the no-makeup image and added an adversarial loss between the warped image and the model's output. As a first stage pairedCycleGAN trains only the makeup removal model as a symmetric domain transfer application and subsequently applies the asymmetric loss for makeup application. While these processes are displayed as two stages, they do not resemble our two staged method since these stages are heavily dependent on each other making it nontrivial to derive a makeup asset using this method. PairedCycleGAN divided the face into patches and trained a makeup transfer model for each patch separately. This approach results in high-quality outputs but suffers from two main disadvantages. First, this method struggles with unknown or "extreme" makeup styles. As the images are processed in patches, it is bound to perform sub-optimally on makeup styles in which the makeup crosses the boundary of the patches, as in many more extreme styles. Second, the makeup extraction process is performed simultaneously with the application, since the joint training is used to train the makeup application network. This makes it impossible to correct extraction errors or select only a part of the makeup and combine it with makeup from other reference images, allowing less flexibility and variety for the user and perhaps also potentially reducing outputs quality. 

Gu \textit{et al.} \cite{gu2019ladn} presented LADN (Local Adversarial Disentangling Network for Facial Makeup and De-Makeup), displaying results for styles including dramatic and high detail styles across multiple facial features. Similarly to Chang \textit{et al.} in pairedCycleGAN, makeup removal and application are trained simultaneously. This method used local discriminators in order to avoid using cropped patches of the images. In contrast to Chang \textit{et al.} this approach used an encoding architecture to disentangle the makeup style and identity encodings, as is also done in \cite{horita2020slgan} . As a result, it was possible to use the codes to produce interpolated makeup styles by interpolating between two style encodings.

Recently, Jiang \textit{et al.} \cite{jiang2020psgan} attempted to tackle this problem using multiple interdependent stages. In the first stage they extracted the makeup style's feature representation using an autoencoder and integrated a warping module into the encoding stage making the method less susceptible to large pose changes. In the second stage they used the feature representations extracted from the first stage, concatenated with the warp features, and used them to decode that output image. This method shows an improvement in the flexibility given to the user, permitting them to combine styles from up to three reference images and controlling the level of makeup that is added. This is achieved through the separation of the faces into facial areas (eyes,lips,skin) as was done in most of the previous methods. Thus, using Jiang \textit{et al.}'s method, it is not possible to combine makeup styles that pass between or within the specified facial segments. For example, using this method one can't simply add two different drawings to the same cheek. It is noted also that none of the examples given by this method included drawings or even multicolored eye shadow, so it has not shown successful results on complex styles.

\subsection{Weakly supervised segmentation}
MakeupBag addresses makeup extraction as a weakly-supervised segmentation task. Weakly-supervised segmentation attempts to remove the dependency of segmentation models on pixel labels, which are expensive to collect. Most frequently, the supervision that exists for this task are class labels of the entire image, but not the specific pixel locations. Several methods have explored object localization using manipulations such as: masking out parts of the image \cite{bazzani2016self}, combining multiple instance learning with deep features \cite{cinbis2016weakly}, using internal features of a classification model \cite{zhou2016learning,hong2017weakly}. Brendel \etal \cite{brendel2019approximating} did not specifically set out to solve this task but our proposed method benefits from their contribution to our segmentation task. Introducing BagNets, they proposed a method of visualizing the prediction of the model and the local features that promote the prediction. In their paper they presented a deep architecture that ensures that patches are trained independently, thus enabling to identify the locations in the image that lead to the model's class prediction. Their surprising result was that even using such limited context, this architecture outputs comparable results to other architectures that use much wider receptive fields.

\section{MakeupBag}

We present MakeupBag, a novel method consisting of two distinct stages: i) makeup extraction ii) makeup application. This approach is designed to empower the user to apply also makeup styles that vary from that of the reference image. 

\subsection{Makeup Extraction}

Our objective in this phase is to produce a makeup asset that is intuitively comprehensible to the user and presents the exact makeup style and locations that will be applied. This allows the user to blend multiple styles, remove a part of an existing style or simply correct imperfections in the the output of the extraction module. This contrasts with current methods that do not produce an editable makeup asset but merely encode the reference makeup style. 

Due to the lack of labelled training data for makeup segmentation, we investigate unsupervised and weakly-supervised approaches.

\subsubsection{Naive methods}

We first consider several simple methods for makeup detection (see appendices for more details): 

\textbf{GMM:} One naive method for makeup extraction is to model the color distribution of skin pixels using Gaussian Mixture Models, and label pixels with low probability under the probability distribution as makeup. 

\textbf{Chroma-deviations:} Another simple method transfers the image to the HSV color space and detects pixels that deviate from the standard skin chroma values.

\textbf{Unsupervised image translation residuals:} The last naive method consists of two stages: i) learning makeup to no-makeup image translation on patches of the face (eyes, lips, skin) using CycleGAN. By the end of training, the generator is used to map each face patch into its no-makeup version. The no-makeup patches are blended back together using Poisson blending to form a no-makeup face. ii) The residual between the original image and its version translated to no-makeup is computed, where pixels with large difference are classified as containing makeup.

\subsubsection{Proposed Extraction}

Inspired by the works of Brendel \etal \cite{brendel2019approximating}, we train a makeup classifier to derive a prediction for the makeup segmentation mask. We manually divide our dataset of facial images into two subdomains: $\cal{X}$, faces with no makeup, and $\cal{Y}$, faces with makeup. The classifier model produces a probability for each patch of $17 \times 17$ pixels which we use as a confidence level for the existence of makeup in that patch. This model is constrained only to local features, thus simplifying the generation of a probability score for specific patches, in contrast to other classification models that can be influenced by global features biasing the scores. As the patches' predictions are independent, we can process each one individually. We wish to preserve the independence and locality of the patches, while utilizing global prior knowledge of typical makeup characteristics. We add to our image input an indicator layer for each facial segment. Different facial parts typically contain corresponding types of makeup with varying color distributions and textures. For example, makeup around the lip area generally appears as shades of pink or red, while around the eyes darker colors can be found (due to the eyeliner or mascara). We encode these global features and feed them to the segmentation model. These features were produced using detected facial landmarks and skin segmentation.

\textbf{Extraction Loss:} Let $\phi$ be the makeup extraction function and let $P_i(y_{ref})$ be patch $i, i= 1, \ldots N_p$ of size 17x17 in reference image $W(y_{ref})$, $y_{ref} \in \cal{Y}$. $W$ represents the warp transformation of $y_{ref}$ to $x_{tar} \in \cal{X}$, the target image to which we would like to transfer the makeup. The warp function is obtained from 2D facial landmarks, causing some artifacts for large pose transformations. The classification prediction of $\phi$ is the average of its predictions on all the patches, $P_i$, in accordance with the independence of the patches:
\begin{equation}\label{m_bar}
    \bar{m} = \frac{1}{N_p} \sum_{i=1}^{ N_p}{\phi (P_i(y_{ref})}
\end{equation}
With $\bar{m}$ as defined in Equation \ref{m_bar} we present our extraction loss with $l$ as the ground truth classification for $y_{ref}$, $\mathcal{L}_{mask}$:
\begin{equation}\label{loss_mask}
    \mathcal{L}_{mask} = l*\log{\sigma (\bar{m})} + (1-l) * \log{{\sigma (1-\bar{m} )}}
\end{equation}
wherein $l = 0$ for $x \in \cal{X}$ and $l=1$ for $y \in \cal{Y}$ and $\sigma$ representing the sigmoid function.

At inference we run the convolutional extraction function $\phi$ on the image $x$ to obtain a makeup mask, $M = \phi(x)$, portraying the confidence score of the classification model that each pixel $M_{i,j}$ contains makeup. We hypothesize that areas that have higher confidence scores for makeup contain stronger makeup, while lower scores correspond to areas that have less makeup or makeup more resembling natural skin.

We present a numerical evaluation which we conducted showing the accuracy of our segmentation model using a test dataset of $150$ images that were hand-labelled for makeup. In Fig.~\ref{fig:roc_curve} we show the accuracy of our method versus the alternatives. See Sec.~\ref{results} for more results.

\begin{figure*}[t]
        {\includegraphics[width=\textwidth]
        {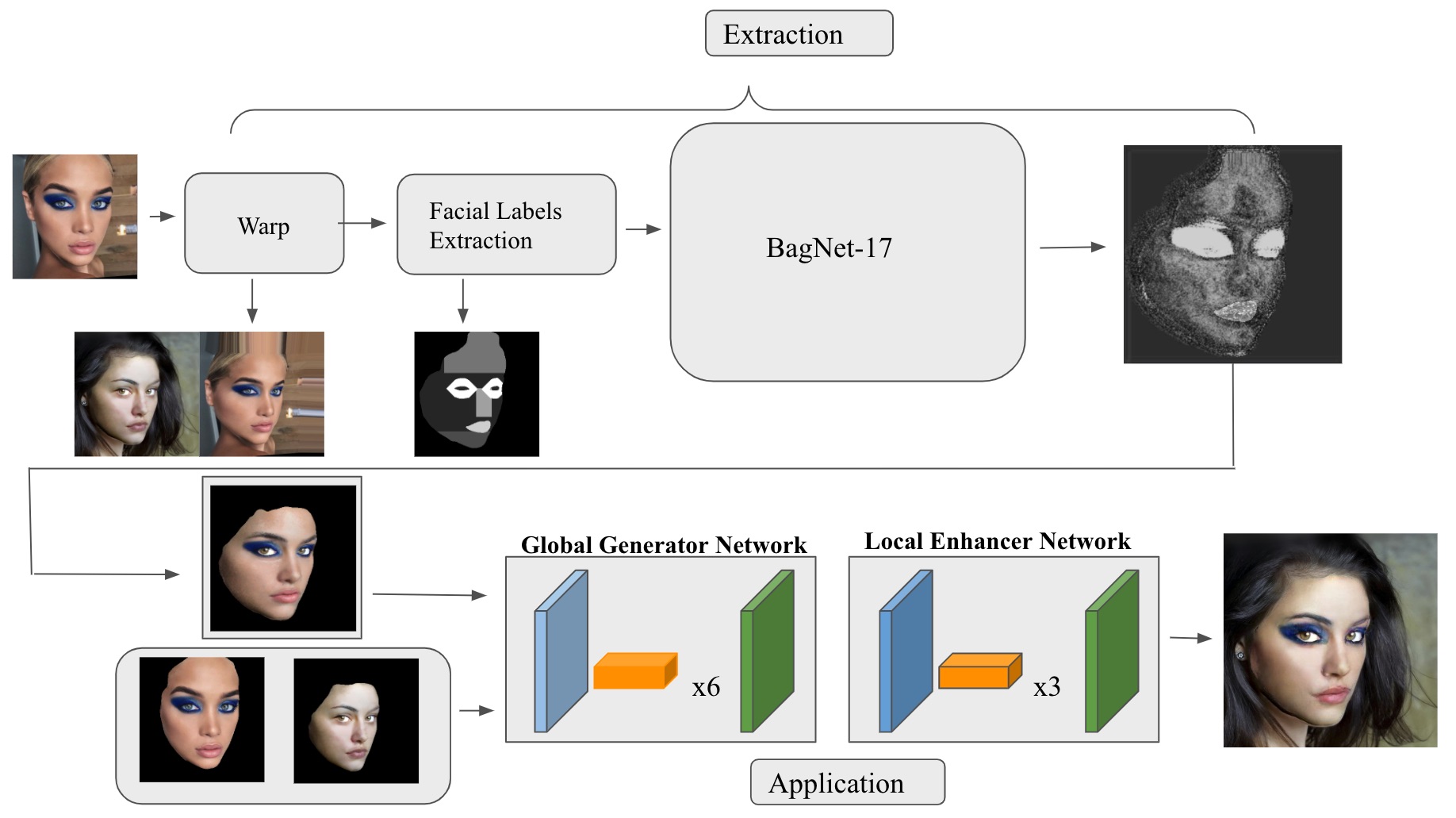}
        \caption{\label{fig:overview_figure} MakeupBag Method Summary. The method can be separated into two stages, extraction and application, although it is not necessary. In the extraction phase, the makeup reference image is aligned to the input image and a facial attributes encoding mask is produced. After this, using a BagNet type model, that is restricted by its architecture to local features only, a makeup segmentation mask is produced. The segmentation mask is used to create an estimated image of the target image wearing the reference style, that is used as ground truth for the L1 loss in the application module. The generator in the application method is made up of the Global Generator and Local Enhancer. Each are composed of convolutional front end (blue blocks), transposed convolutional back end (green blocks) and residual blocks (orange blocks). There are 6 residual blocks in the global generator and 3 blocks in the local enhancer. }}
\end{figure*}

\subsection{Makeup Application}

Makeup segmentation extraction by itself is insufficient for producing high-quality transfer for more challenging styles i.e. simply adding the mask using a linear combination of the target and reference image can generate imperfect results in harder cases. This task faces several challenges: i) the makeup mask can have jagged edges that lack a natural shape ii) the mask has strong gradients at the edges producing an unrealistic look iii) in many cases there is a difference in texture between faces with and without makeup that does not need to be transferred, as the texture of the reference image is not a specific attribute of the reference image, but rather in many cases, due to a base makeup layer that is applied to the entire face. We would like the artificial makeup image to possess this quality as well in areas that have natural looking skin. Our goal is to learn a function that takes as input the no-makeup and reference images as well as a segmentation map and outputs a realistic image $I_{est}$ with the transferred makeup. In other words, we would like $I_{ext}$ to satisfy: $I_{est} \in \cal{Y}$ as well as conserving the identity of $x_{tar}$.  

The key for the success of our method is utilizing the mask extracted in the previous stage, which allows us to use supervised image translation techniques (even though we are in the \textbf{unsupervised} setting). Let $\cal{Y}$ be the domain of faces with makeup and $\cal{X}$ be the non-makeup domain. Our task is to train a generator that maps from the no-makeup domain $\cal{X}$ to the makeup domain $\cal{Y}$, $G: {\cal{X}} \rightarrow {\cal{Y}}$. This can be formally written as:
\begin{equation}
    I_{est} = G(x_{est}, M, y_{ref})
\end{equation}
In order to learn the appropriate makeup transfer function, we propose several loss functions encoding our requirement from the makeup transfer function.

 \textbf{Segmentation Reconstruction Loss:} In the previous stage, we computed a segmentation mask. This loss term penalizes deviations from the segmentation. We encourage pixels to which our model gave a low probability of containing makeup, to resemble those in the target image $x_{tar}$, while encouraging pixels with a high probability of makeup to resemble the reference image $y_{ref}$. This is formally written as follows:
 \begin{equation}
    \begin{split}
        \mathcal{L}_{rec} = & \|M \odot I_{ref} - M \odot I_{est}\|_1 + \\
        &\|(1-M) \odot I_{orig} - (1-M) \odot I_{est}\|_1
    \end{split}
 \end{equation}

 \textbf{Adversarial Losses:} 
We would like to encourage the generator $G$ to output images resembling those in domain $Y$. We train a discriminator network, $D$, to attempt to classify whether images are produced by $G$ or whether they originate from the ${\cal{Y}}$ domain.  At the same time, we train $G$ to try to fool the discriminator $D$ and generate images that are indistinguishable with those of ${\cal{Y}}$. Upon successful training, the distribution of images produced by $G$ looks exactly like ${\cal{Y}}$ and no discriminator can distinguish between them (in practice such convergence is only partially achieved). The loss function for training the two networks is given by: 
 \begin{equation}
         \mathcal{L}_{GAN}(G,D) = (D(G(I_{est},M, y_{ref}))-1)^2 + D(y_{ref})^2
\end{equation}
 
 \textbf{Overall Loss: }
 In our method we aim to reach an optimum of the following min-max problem:
 \begin{equation}
     G', D', \psi' = \arg \min_{G, \phi} \max_{D} {\mathcal{L}_{mask}  + \lambda_1 \mathcal{L}_{rec} + \lambda_2 \mathcal{L}_{GAN}}
 \end{equation}
where $\lambda_1$ and $\lambda_2$ are hyper parameters of the model which we use in order to increase our control on the discriminator loss as necessary. In practice, we used the following relative factors between the different loss terms: $\lambda_1 = 40, \lambda_2 = 1 $.

\begin{figure*}
\begin{center}
%\begin{tabular}{cccccccc} 
\includegraphics[width = 0.11\textwidth]{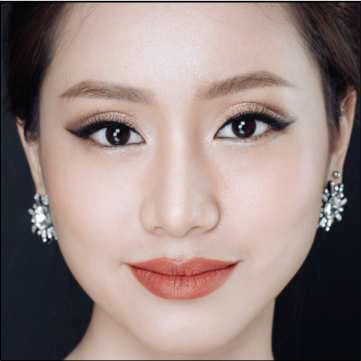}
\includegraphics[width = 0.11\textwidth]{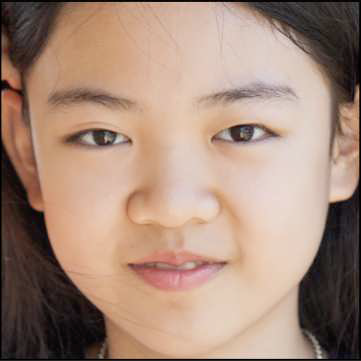}
\includegraphics[width = 0.11\textwidth]{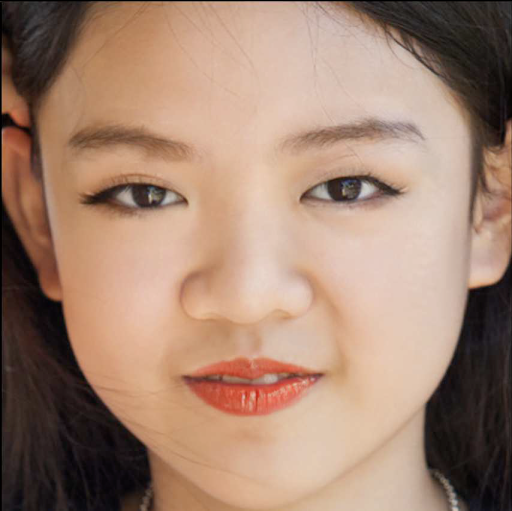}
\includegraphics[width = 0.11\textwidth]{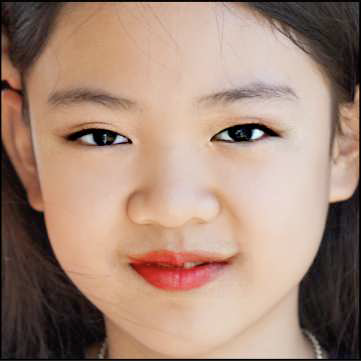}
\includegraphics[width = 0.11\textwidth]{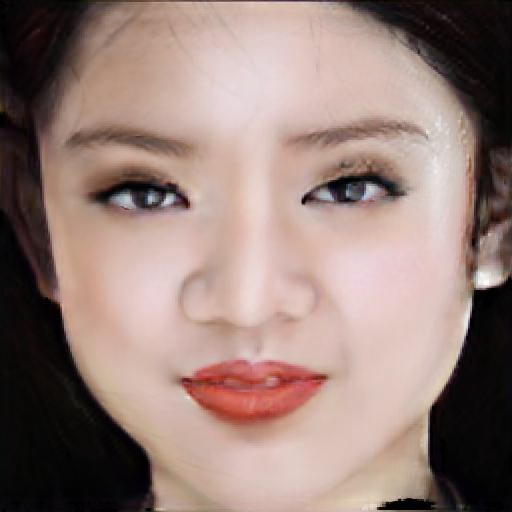}
\includegraphics[width = 0.11\textwidth]{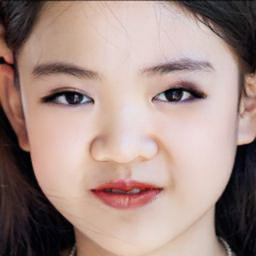}
\includegraphics[width = 0.11\textwidth]{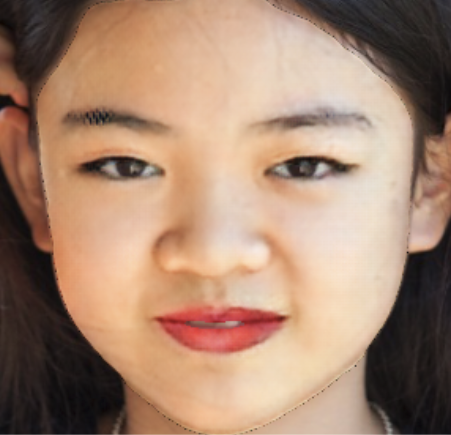}
\includegraphics[width =  0.11\textwidth]{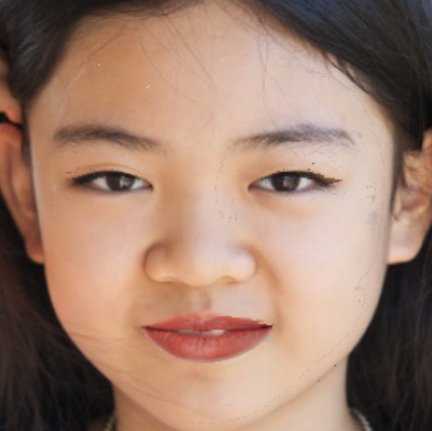}\\
\includegraphics[width = 0.11\textwidth]{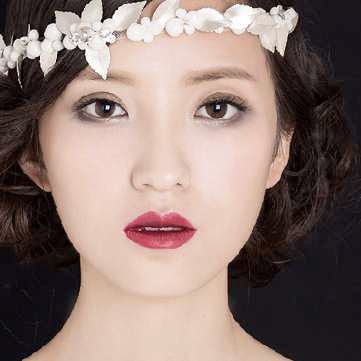} 
\includegraphics[width = 0.11\textwidth]{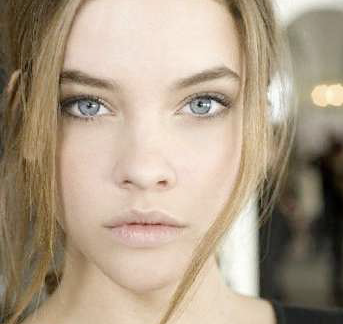}
\includegraphics[width = 0.11\textwidth]{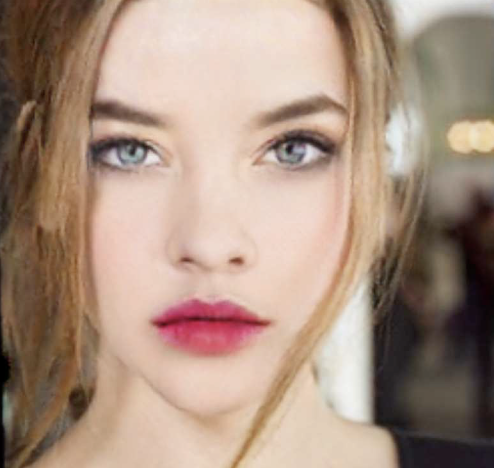}
\includegraphics[width = 0.11\textwidth]{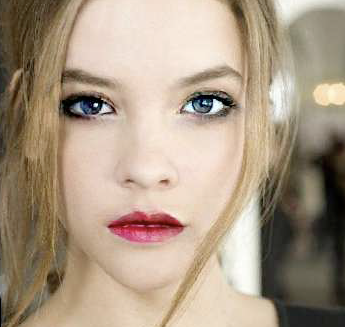}
\includegraphics[width = 0.11\textwidth]{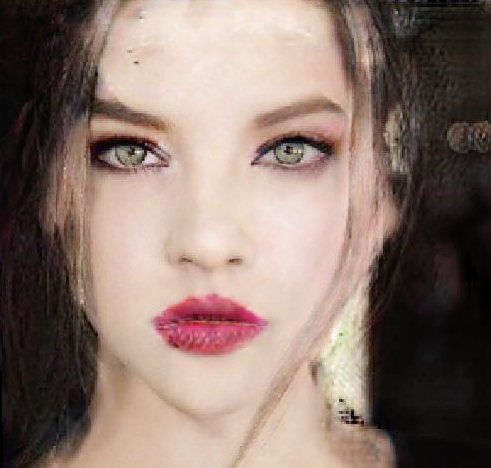} 
\includegraphics[width = 0.11\textwidth]{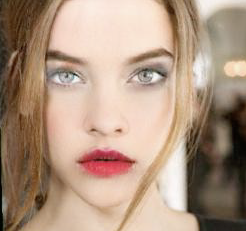}
\includegraphics[width = 0.11\textwidth]{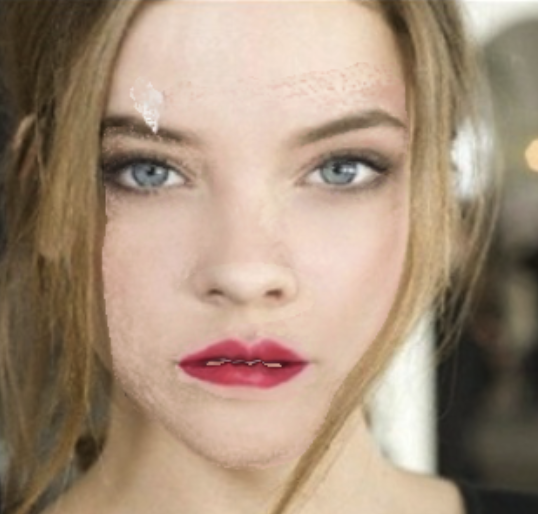}
\includegraphics[width =  0.11\textwidth]{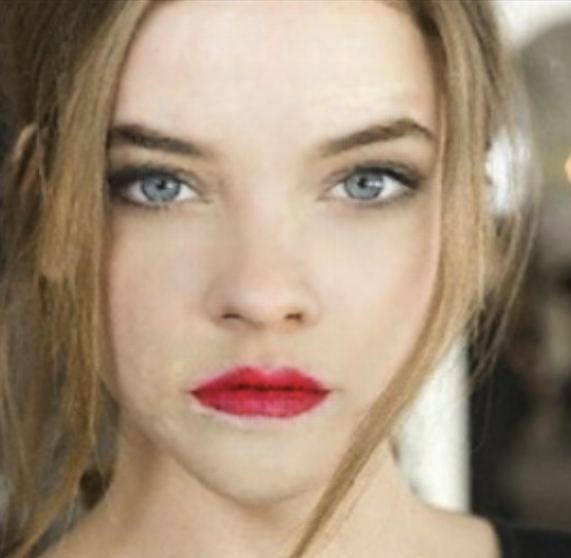}\\

reference ~~~~~~~~ target ~~~~~~~~~~~ Paired ~~~~~~~ BeautyGlow ~~~~~ LADN ~~~~~~~~~ PSGAN ~~~~~~~~~~~ Ours ~~~~~~~~~~~~ Ours~~~~~ \\
 ~~~~~~~~~~~~~~~~~~~~~~~~~~~~~~~~~~~~~~~~CycleGAN ~~~~~~~~~~~~~~~~~~~~~~~~~~~~~~~~~~~~~~~~~~~~~~~~~~~~~~~~~~~~~~~~~~~~~~~~~~ 1st stage ~~~~~~~~~ Final

\end{center}

%\end{tabular}
\caption{Qualitative Comparison to Baseline Methods: we observe the  predictions of MakeupBag in comparison to many of its baselines. Some of our competitor's predicted images do not preserve the wearer's identity. For example, PSGAN, in both examples, does not completely conserve the wearers identity: in the second row, due to the asymmetric eyes, and in the first row since the skin has fully transformed to the reference image's shade. We provided the outputs of the extraction phase as well as the final result (which includes the GAN based application model). We see that the second phase improves a lot on the first one. Overall, MakeupBag fully preserves identity and imitates the reference's style with high accuracy.}
\label{fig:baselines_comp}
\end{figure*}
\begin{figure*}

\begin{tabular}{c} 
Reference~~~~~~~~~~~~~~~~~~~Target~~~~~~~~~~~~~~~~~~~BeautyGAN~~~~~~~~~~~~~~~~~~~LADN~~~~~~~~~~~~~~~~~~~PSGAN~~~~~~~~~~~~~~~~~~~MakeupBag\\
\includegraphics[width = \textwidth]{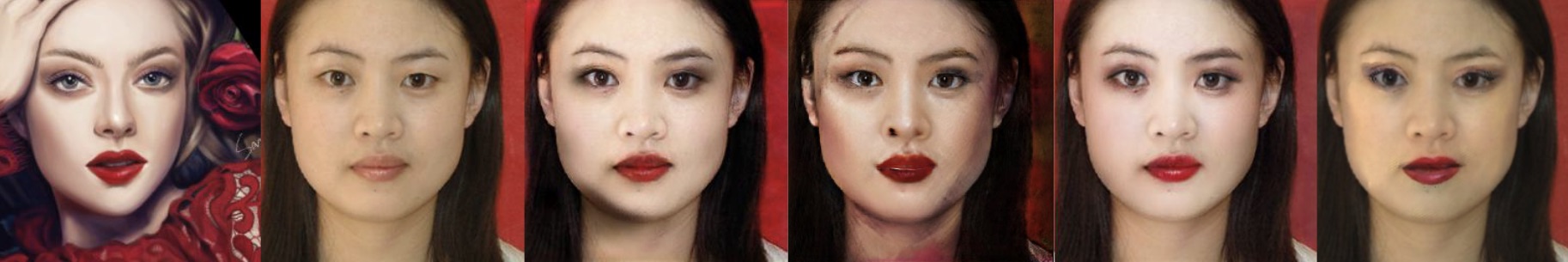}\\
\includegraphics[width = \textwidth]{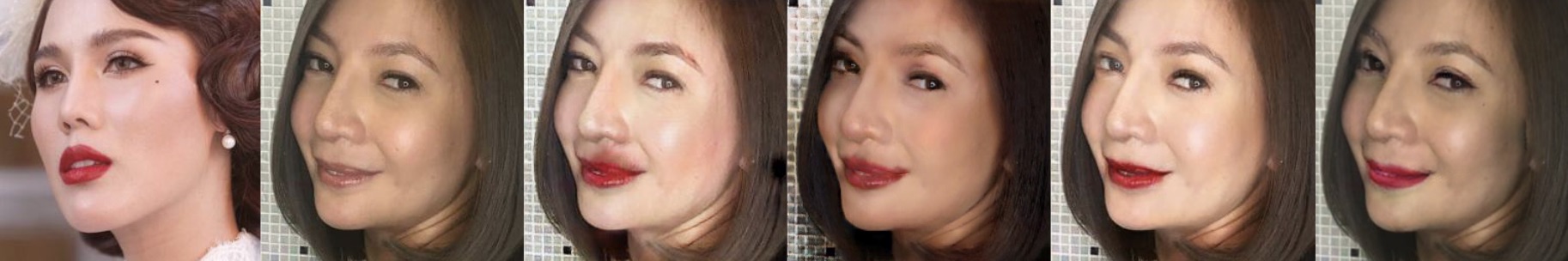}\\
\includegraphics[width = \textwidth]{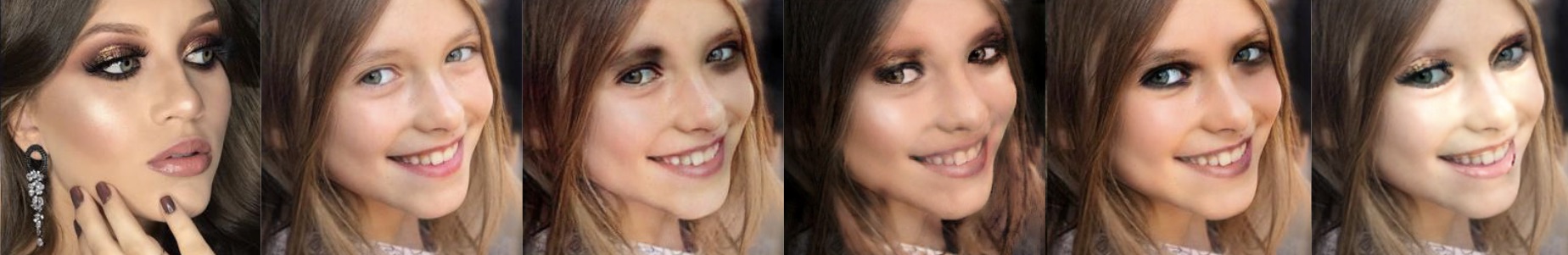}\\
\includegraphics[width = \textwidth]{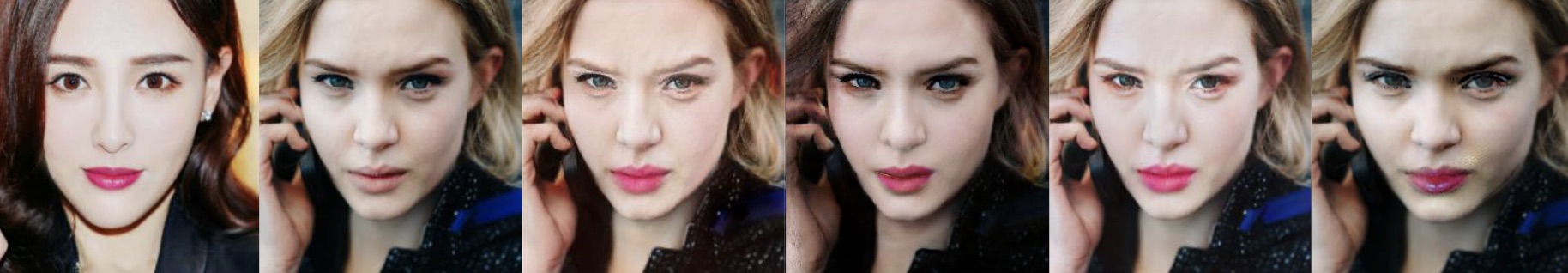}\\
\includegraphics[width = \textwidth]{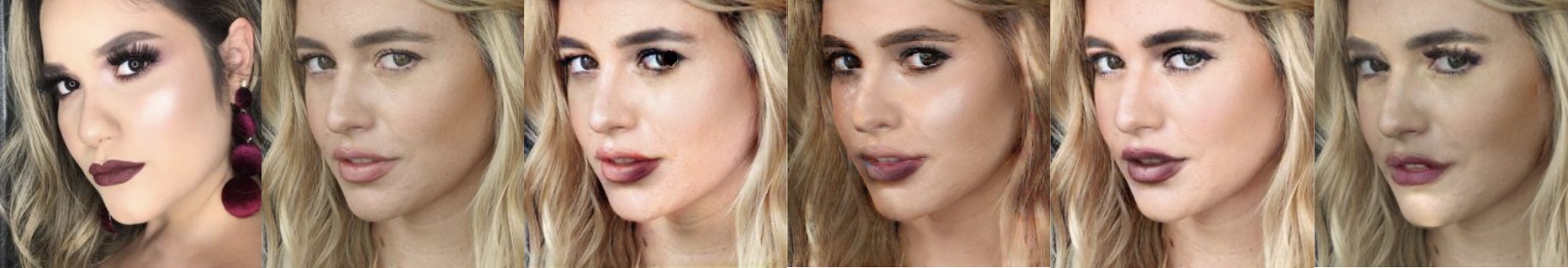}\\

\end{tabular}
\caption{Qualitative Comparison to Baseline Methods (unusual poses and expressions): We show comparisons to three of our baselines; BeautyGAN, LADN and PSGAN. Our method produces outputs that both conserves the wearers identity (not altering skin color substantially as is apparent in some of the other methods) and clearly transfers similar makeup styles to the reference images.  }
\label{fig:baselines_comp2}
\end{figure*}
\textit{Limitations:} We note that applied makeup that appears identical to the natural skin would not be transferred, as our method would not be able to detect that the skin has changed from its natural state. We do not believe this is a significant limitation, and could even be considered an attribute of the method, as we want to preserve identity and changing the color of skin in $x_{tar}$ would reduce the likeness of the input and output images.  A simple mitigation is to modify the makeup color manually by a constant offset before the makeup application stage. Occluding objects (glasses, microphones etc.) are typically classified as makeup, as they do not appear to be natural skin. Transferring such objects is however not the desired behavior. 

\subsection{Implementation Details}
The architecture for the extraction model is based upon Resnet-50 with small alterations of replacing most of the $3 \times 3$ convolution kernels with $1 \times 1$ convolutions to reduce the receptive field (thus confining the model to local features). The last layer is a fully connected layer with an output of $1$, which sums over the feature vector that was output from the model. We add a sigmoid activation layer, producing the probability, $p$, predicting whether the overall image belongs to domain $\cal{X}$ or $\cal{Y}$. We optimize against a cross entropy loss function as seen in Equation \ref{loss_mask}. 

For our generator architecture we use a coarse to fine architecture similar to that of pix2pixHD, which has proven success in generating high quality images of up to 2048x1024 pixels. It consists of a global generator and local enhancer network, where the local enhancer outputs an image with a factor 4 higher resolution than the input it was given. The global generator consists of 6 residual blocks with a convolutional front-end and a back-end of transpose convolutions. The local enhancer network is also composed of 3 residual blocks and a convolutional front end and transposed convolutional back-end. The difference between the global and local generators is that the input to the residual blocks is the sum of the feature map of the output of the global generator and the feature map that comes out of the front end of local enhancer. 

Pix2PixHD uses a conditional GAN loss which does not suit our task, we therefore use an unconditional GAN discriminator. The implementation that we employ is the same as that of MUNIT,  who utilizes a multiscale discriminator composed of convolutional blocks. 

We optimized our model using SGD with ADAM optimizer with learning rate of 0.0001 for the discriminator and 0.0002 for the generator, $\beta _1 = 0.5$, $\beta _2 = 0.99$ for both. We trained up to 50 epochs until visually satisfying results were achieved. 

While there is no prevention in training the the entire method end to end (while still preserving the option of editing the makeup segmentation mask), as is seen in our loss formulations, we trained them separately and then ran them one after the other, due to the initial instability of the GAN model training.

\section{Experiments and Discussion }

In this section we will showcase some of our model's output images; we will show how the makeup can be controlled and how the agility of the method allows us to combine multiple styles in one look. We will compare our results to our current baseline on images that previous research have presented in their works and show that our method improves on them. 
\subsection{Dataset}

For both the segmentation training and the makeup application model we used datasets we manually labelled as with and without makeup and in some cases we also collected the images (as there was a lack of extreme makeup styles in the existing datasets).  We made use of  CelebA and FFHQ datasets after labelling them into "with makeup" and "without makeup", but most images have very mild and conservative makeup which is insufficient for our purposes. Consequently, we additionally manually downloaded and labelled images from the web, images that express a greater diversity in makeup styles, some more extreme. Altogether we collected approximately 4000 images of size 1024x1024 (for each category) for the purpose of training our model. Generally, there are many more faces with light makeup than faces with mid-level to extreme makeup. In order to neutralize the bias effect that this may cause particularly in adversarial training, we oversampled heavier makeup styles. We also added data augmentation to our data by flipping our images horizontally. These changes proved crucial to the model's successful training.

\subsection{Results}\label{results}

We evaluate the results of makeup extraction and application separately.

\subsubsection{Makeup Extraction}

\begin{figure}[!ht]
        \includegraphics[scale=0.5]
        {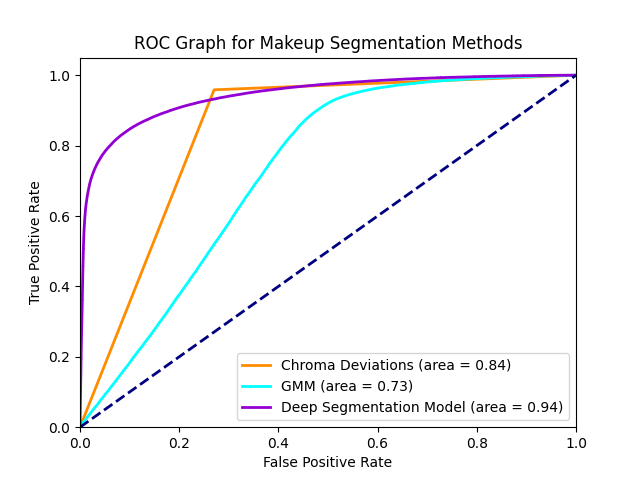}
        \caption{\label{fig:roc_curve} Numerical comparison between the extraction methods: From the ROC curve and the AUC we can see that our deep extraction method significantly outperforms the two others.  }
\end{figure}

In the method overview section, we presented two alternatives to our weakly supervised segmentation mask: GMM and Chroma-deviations. We first created manually segmented dataset of 
approximately $150$ makeup images. We compared the accuracy of the different segmentation methods against the groundtruth labels. We plotted the Receiver Operating Characteristic (ROC) curve (Fig. \ref{fig:roc_curve}) of the three methods and the area the under the curve (ROC AUC. It is apparent from the graph and the AUC that while the other two methods (particularly "chroma deviations") provide a credible mask, MakeupBag outperforms, with an AUC of 94\%. This provides evidence that MakeupBag is a better makeup segmentation method than the other two.

\subsubsection{Makeup Application}

For comparison with other methods, we evaluated our method on the dataset introduced by Gu \textit{et al. }, containing images used in previous works. The images in the LADN dataset are of lower resolution than the resolution that our model was originally trained on. Nevertheless, to ensure that we can conduct a fair comparison, we trained our model on exactly the same images used and therefore trained a second, identical model, for lower-resolution images. The application model's architecture resembles the one for large images, although we optimized the hyper-parameters of this model for lower resolution images.

 We provide a visual comparison Fig. \ref{fig:baselines_comp} between the results of MakeupBag and of key baselines. In the top row, we can see that  several competing methods made the entire color of the skin whiter (LADN, PSGAN). This is not a desirable outcome of the method as people trying to imitate a makeup style do not generally intend to change the shade of their skin. Methods that modify the color of the skin, do not conserve the wearer's identity. Moreover, PSGAN, in both examples, does not completely conserve the person's identity. In the second row, for example, the output of PSGAN seems to give the eyes an asymmetric look and the style is not fully copied in the eye area. 
 
 To ablate the results of the application networks, we present the outputs of the extraction phase as well as the final result (which includes the GAN based application network). In the second example some of the crown on the forehead was recognized by the extraction method as makeup but we can see that the application method smoothed it over and improved the overall result. Our method copied the makeup seamlessly and most accurately while fully preserving the identity of the target image. 
 
 In Fig. \ref{fig:baselines_comp2} we show comparisons to three of our baselines on unusual and complicated poses and expressions showing that also on these hard cases our method outperforms its competitors and produces convincing makeup transfer while completely preserving the identity of the wearer (for example, our method does not change the skin color of the wearer significantly).
 
\begin{figure}[ht]
        \center{\includegraphics[scale=0.16]
        {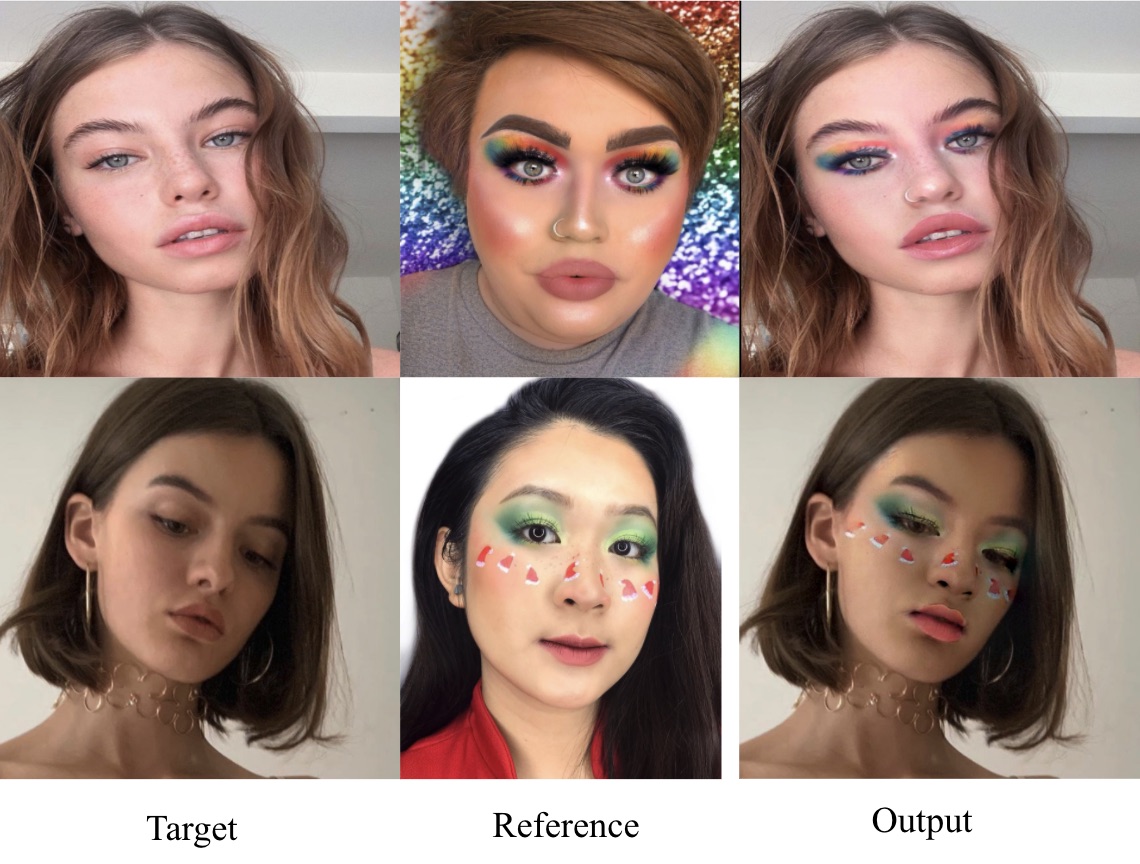}
        \caption{\label{fig:extreme_results} Results on extreme styles. We show high quality results on styles with multiple colors and drawings. The tricky makeup styles are transferred in a realistic way. The minor flaw is that the extraction method recognizes the nose ring as makeup and transfers it. This can be corrected using mask manipulation due to the separability of the phases. }}
\end{figure}

 In Fig. \ref{fig:extreme_results} we show results of extreme and complex makeup styles, including multiple colors and drawings on the face. We show that our model transfers these makeup styles well, the results appear realistic and the identity of the target face is well preserved. A minor flaw that can be seen in row $1$, the nose ring was also extracted as makeup and was therefore transferred.

\begin{figure}

        \center\includegraphics[scale=0.2]
        {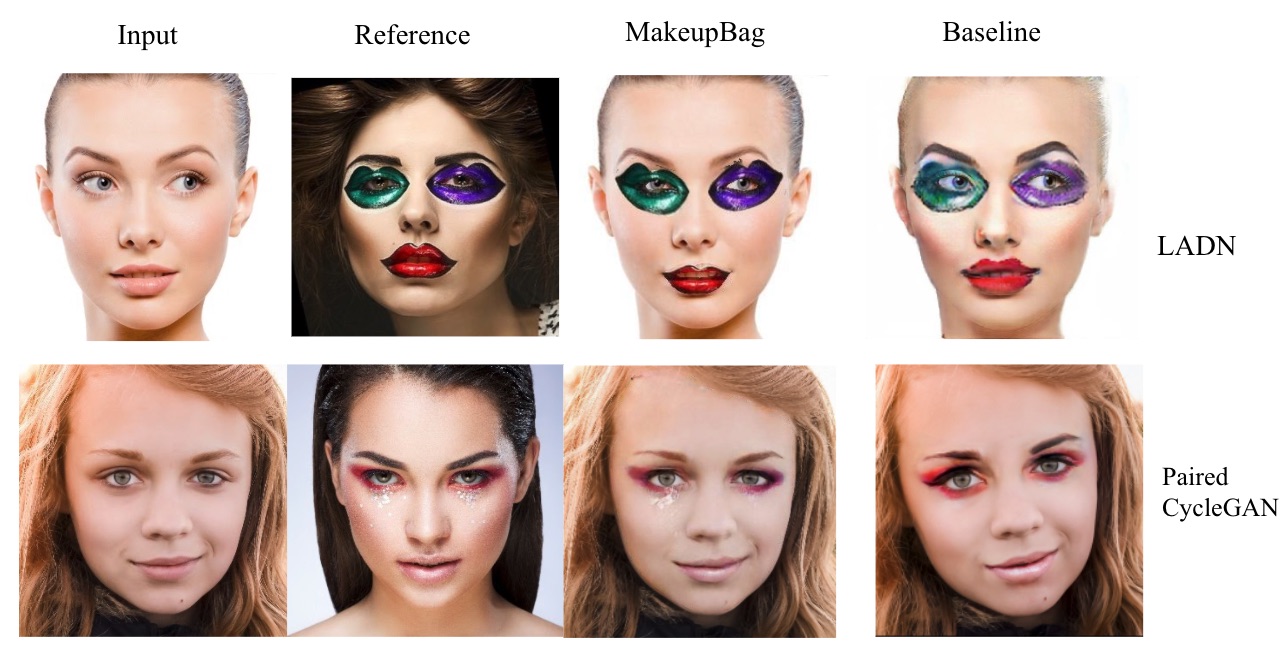}
        \caption{\label{fig:ladn_comp}  Comparison to baselines in difficult cases. We compare our performance to LADN's outputs in the first row and pairedCycleGAN in the second row. We can see that LADN failed to fully preserve the input image's identity as well as losing the black lip liner and some of the color around the eyes while MakeupBag did not have these flaws. The second row was observe a pairedCycleGAN failure case. Indeed, we see that the eyeshadow is much lighter than in the reference and none of the glitter can be seen in the baseline image. We note that MakeupBag transferred most of the glitter and has replicated the eye shadow accurately. This figure demonstrates the superior results of MakeupBag to the baselines in both style transfer and identity preservation.  }
\end{figure}

\textbf{Comparison to LADN and pairedCycleGAN:} In Figure \ref{fig:ladn_comp} we compare our performance to LADN's outputs in the first row and pairedCycleGAN in the second row. We can see that LADN failed to fully preserve the input image's identity as well as losing the black lip liner and some of the color around the eyes while MakeupBag does not have both flaws. The second row shows a fail case brought by pairedCycleGAN in their paper. They claimed that their method cannot transfer the glitter and the color of the eye shadow. It is noted that MakeupBag transferred most of the glitter and has replicated the eye shadow accurately. This figure shows that we have superior results to our baseline in both the style transfer and in identity preservation.

\begin{figure}[ht]
        \center{\includegraphics[width=0.45\textwidth]
        {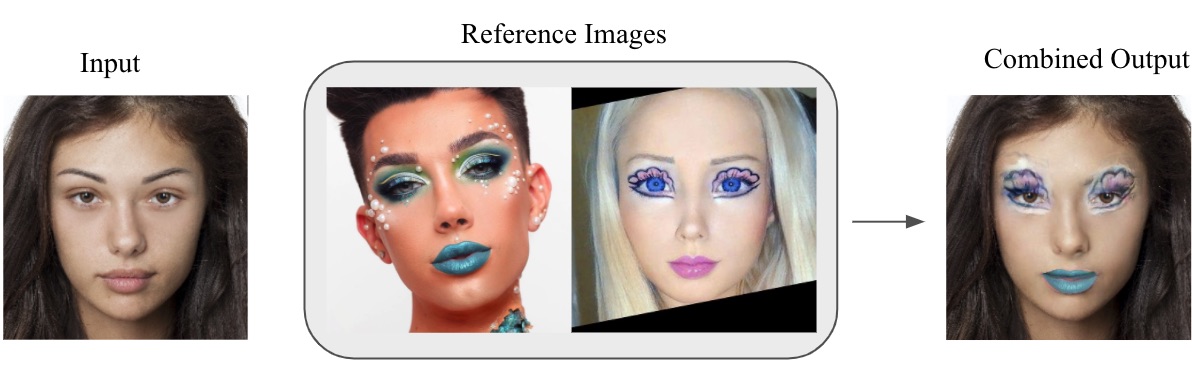}
        \caption{\label{fig:3_ref} Creating one new look from three different reference images. We combine these two extreme styles into one. This enables the creation of diverse makeup styles. }}
\end{figure}

\textbf{Combining Multiple Styles:} One of the main advantages of our method is the ability to edit the result of the makeup extraction before the makeup application and thus create numerous new styles. In Figure \ref{fig:3_ref} we demonstrate this functionality where we use two reference makeup styles to produce one makeup look, one for the eyes and one for the lips. This was possible due to the fact that an asset was produced and so we could remove the lips in one asset and in the other remove the eyes and so we could produce a new look that did not exist initially. The variety of makeup styles that can be created using this method are virtually limitless, by creating combinations of existing makeup styles.

\section{Conclusion and Future Work}
We presented a novel approach for makeup transfer, disentangling the makeup extraction and application tasks. Our approach has been shown to achieve state-of-the-art performance in an extensive experimental analysis. The advantages are both the high-fidelity of makeup transfer as well as allowing the creation of diverse novel makeup styles, through combination of styles. Future work can focus on collecting larger datasets of extreme makeup styles as this is the current bottleneck for learning-based makeup style-transfer methods.

\newpage
{\small
\bibliographystyle{ieee_fullname}
\bibliography{main.bib}
}
\begin{appendices}

\section{Introduction}

We present additional implementation details, segmentation mask examples,  comparisons to other methods and results of our method.

\section{Example of Different Segmentation Masks}

\begin{figure}[!ht]
        % \center{\includegraphics[scale=0.25]
        \includegraphics[scale=0.25]
        {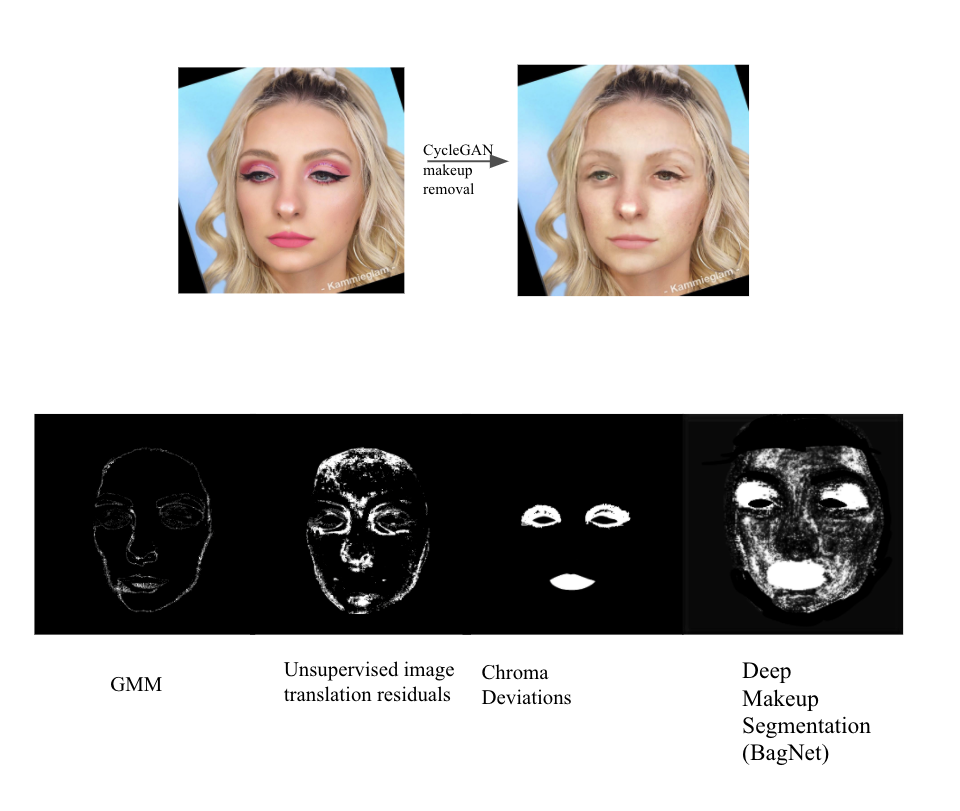}
        \caption{\label{fig:mask_compare} Visual comparison of makeup extraction methods. }
\end{figure}

In the paper we described alternative methods we attempted for makeup extraction. We referred to them as: GMM, Chroma-deviations and Unsupervised image translation residuals. In Figure \ref{fig:mask_compare} we show the output of these masks for one image. We can see that the Chroma Deviations method outputs the most accurate mask (the least noisy) out of the binary masks. BagNet is currently the only one that outputs an alpha mask making the makeup that is also not very noticeable also transferable and improving the overall output. While the mask  has more positive labeling than the mask from Chroma Deviations, these should not be considered as false positives, as the areas that don't have very noticeable makeup receive lower score thus it captures some of the light makeup on the cheeks while still differentiating between light and strong makeup. In the paper we have shown a numerical evaluation demonstrating that the performance of BagNet is the most accurate, while the second most accurate are the masks produced by Chroma Deviations.

\section{Makeup Segmentation Model Architecture}

\begin{figure}
        % \center{\includegraphics[scale=0.25]
        \includegraphics[scale=0.9]
        {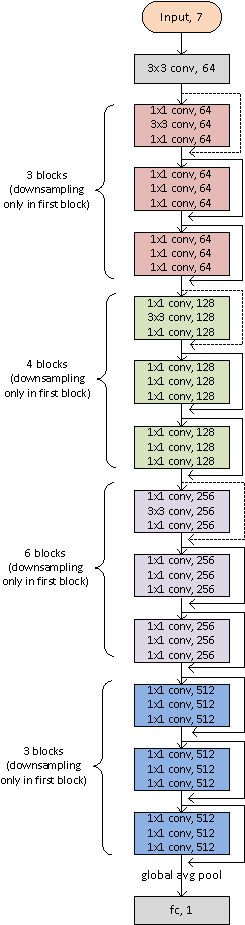}
        \caption{\label{fig:arch} Diagram of MakeupBag's Extraction Stage. }
\end{figure}

As was explained in the paper, MakeupBag's first stage (the extraction stage) is implemented using BagNet-17 \cite{brendel2019approximating} architecture with differences only in the input and output layers. In Figure. \ref{fig:arch} we show a diagram of the model's architecture. Note that it is very similar to the architecture of ResNet-50 with minor changes to the kernel sizes that are configured to oblige the model to use local features only. Our inputs include 7 channels instead of 3 as we add the facial area label masks.

\section{More Output Examples}
\begin{figure*}[t]
\begin{center}
\begin{tabular}{c}

Input~~~~~~~~~~~~~~~~~~~~~~~~~~~~~~Reference~~~~~~~~~~~~~~~~~~~~~~~~~~~~~~~~Output \\
\midrule
\includegraphics[width=0.8\linewidth]{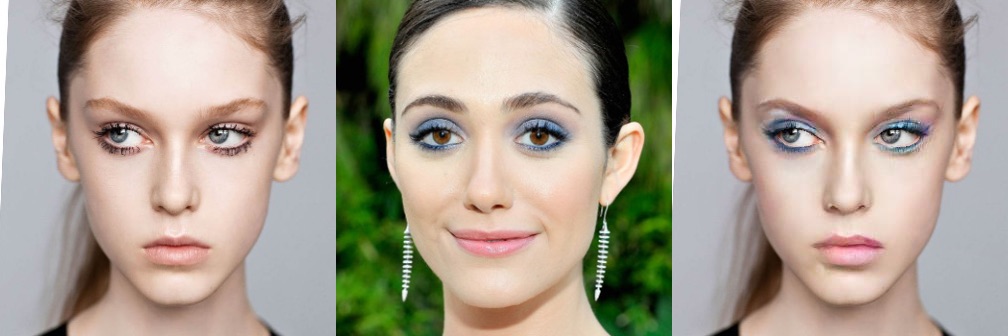}\\

\midrule
\includegraphics[width=0.8\linewidth]{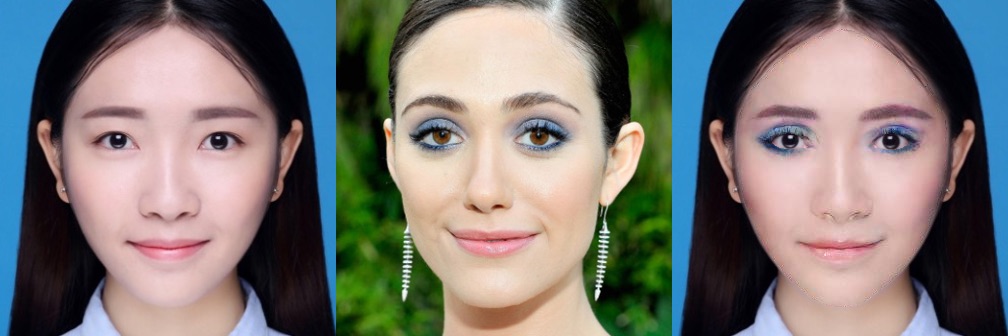}\\

\midrule
\includegraphics[width=0.8\linewidth]{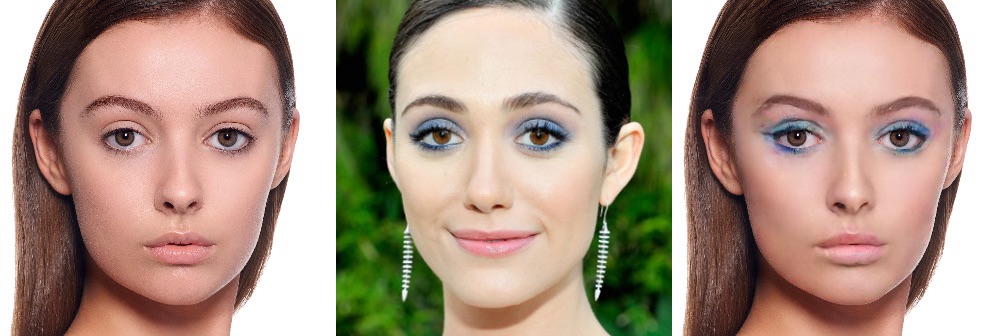}\\

\midrule
\includegraphics[width=0.8\linewidth]{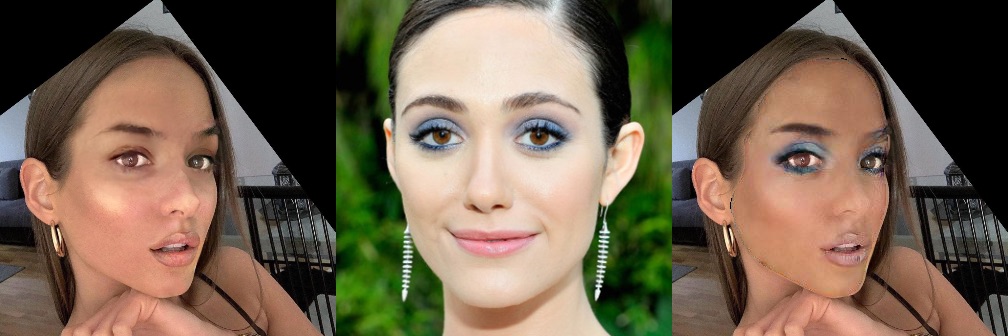}\\

\end{tabular}
\end{center}
\caption{BagNet Output Examples }
\label{fig:app:face1}
\end{figure*}

Further results of the method are presented in Fig. ~\ref{fig:app:face1} - \ref{fig:app:face}. Our method shows good results on various makeup styles and input faces in multiple pose and lighting combinations. Below we show triplets of input image (no makeup face), reference image (makeup style) and MakeupBag output. 

% \newpage

\begin{figure*}[t]
\begin{center}
\begin{tabular}{c}

Input~~~~~~~~~~~~~~~~~~~~~~~~~~~~Reference~~~~~~~~~~~~~~~~~~~~~~~~~~~~~~Output \\

\midrule
\includegraphics[width=0.8\linewidth]{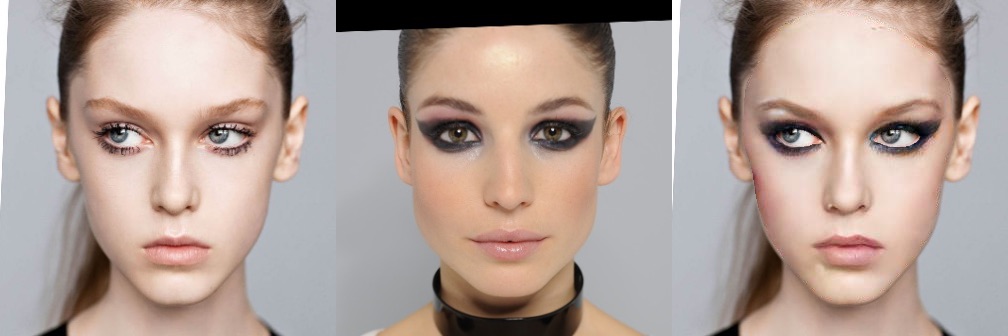}\\

\midrule
\includegraphics[width=0.8\linewidth]{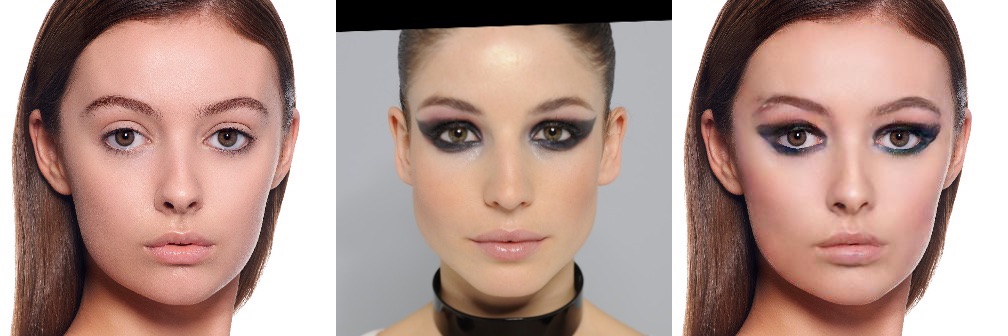}\\

\midrule
\includegraphics[width=0.8\linewidth]{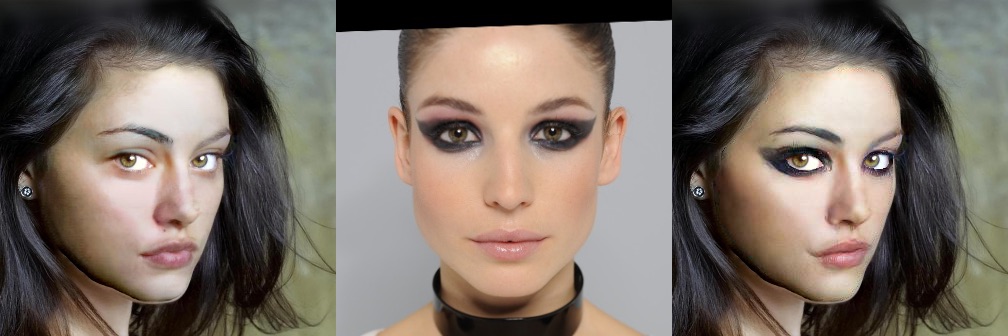}\\

\midrule
\includegraphics[width=0.8\linewidth]{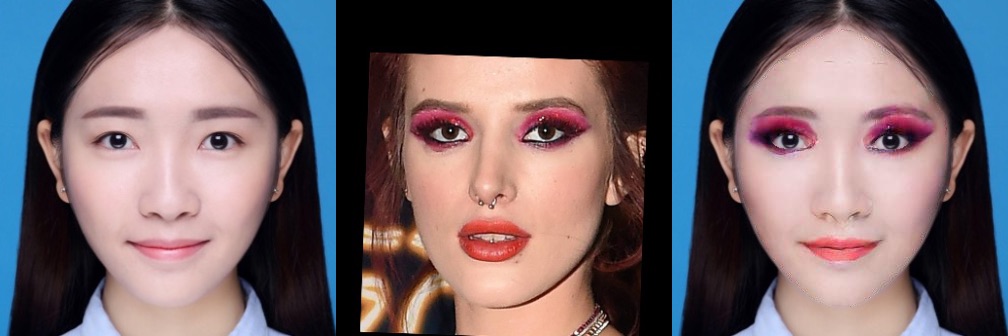}\\

\end{tabular}

\end{center}
\caption{BagNet Output Examples }
\label{fig:app:face}
\end{figure*}

\begin{figure*}[t]
\begin{center}
\begin{tabular}{c}

Input~~~~~~~~~~~~~~~~~~~~~~~~~~~~Reference~~~~~~~~~~~~~~~~~~~~~~~~~~~~~~Output \\

\midrule
\includegraphics[width=0.8\linewidth]{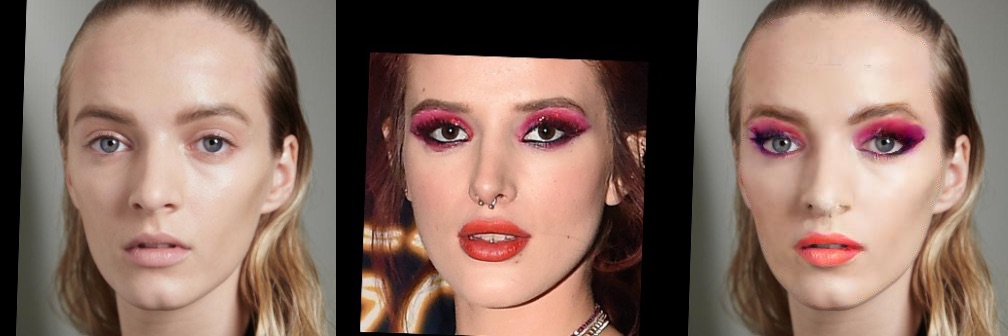}\\

\midrule
\includegraphics[width=0.8\linewidth]{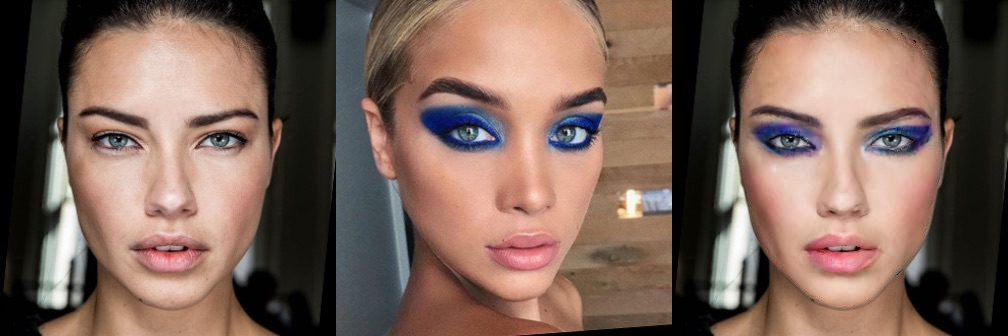}\\

\midrule
\includegraphics[width=0.8\linewidth]{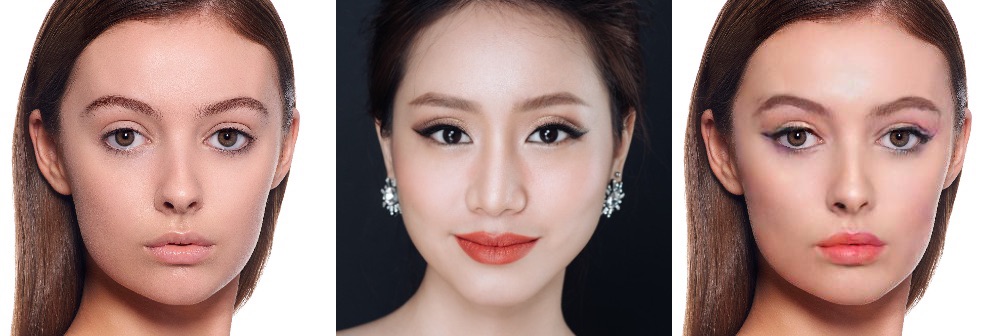}\\

\end{tabular}

\end{center}
\caption{BagNet Output Examples }
\label{fig:app:face}
\end{figure*}

\begin{figure*}[t]
\begin{center}
\begin{tabular}{c}

Input~~~~~~~~~~~~~~~~~~~~~~~~~~~~~~Reference~~~~~~~~~~~~~~~~~~~~~~~~~~~~~~Output \\

\midrule
\includegraphics[width=0.8\linewidth]{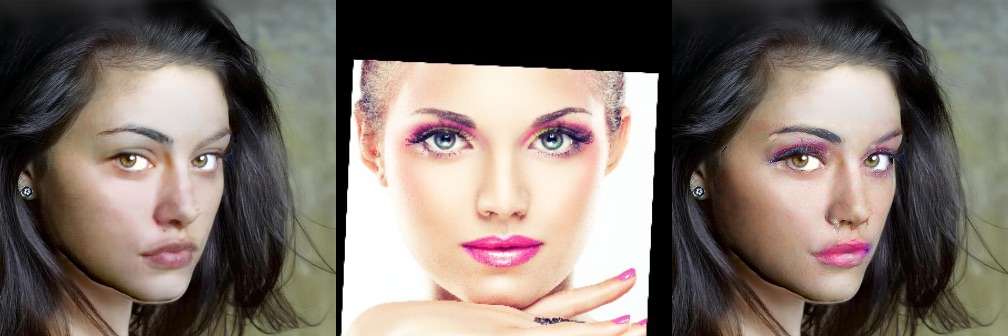}\\

\midrule
\includegraphics[width=0.8\linewidth]{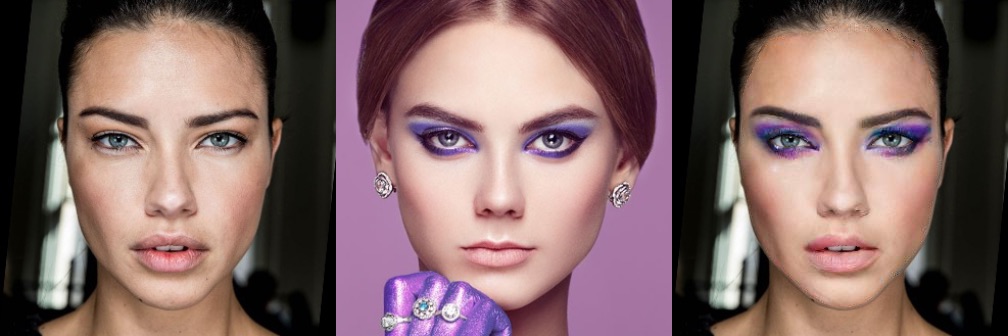}\\

\midrule
\includegraphics[width=0.8\linewidth]{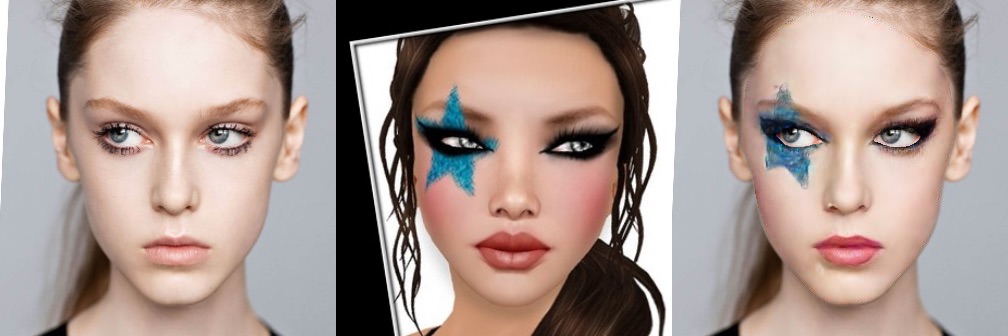}\\

\end{tabular}

\end{center}
\caption{BagNet Output Examples }
\label{fig:app:face}
\end{figure*}

\begin{figure*}[t]
\begin{center}
\begin{tabular}{c}

Input~~~~~~~~~~~~~~~~~~~~~~~~~~~~Reference~~~~~~~~~~~~~~~~~~~~~~~~~~~~~~Output \\

\midrule
\includegraphics[width=0.8\linewidth]{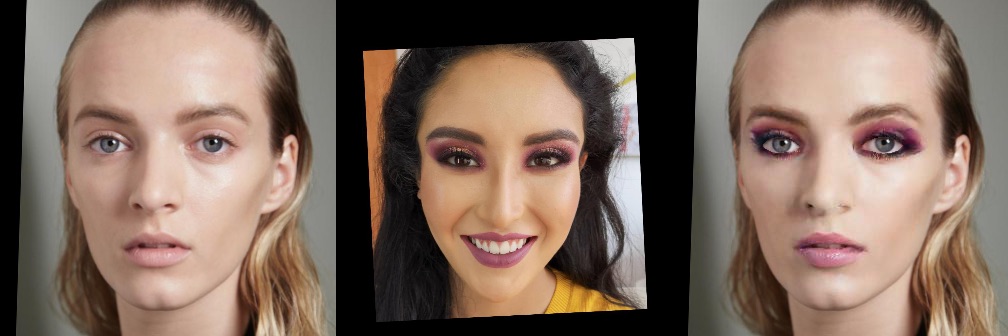}\\

\midrule
\includegraphics[width=0.8\linewidth]{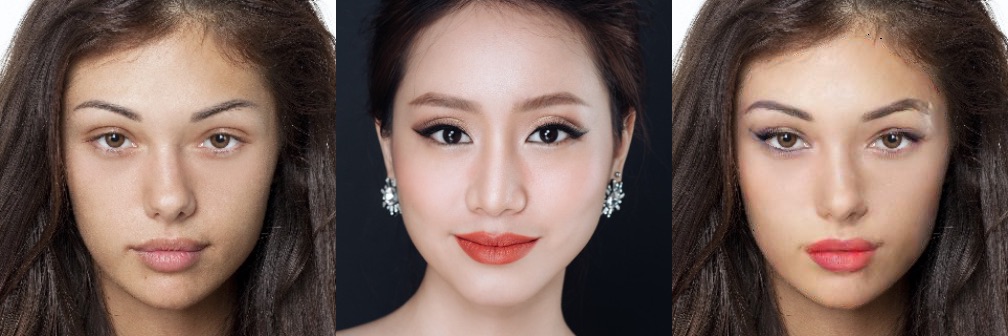}\\

\midrule
\includegraphics[width=0.8\linewidth]{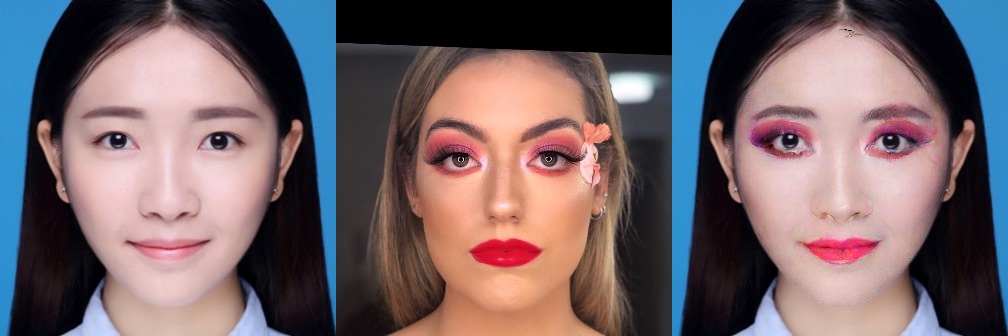}\\

\midrule
\includegraphics[width=0.8\linewidth]{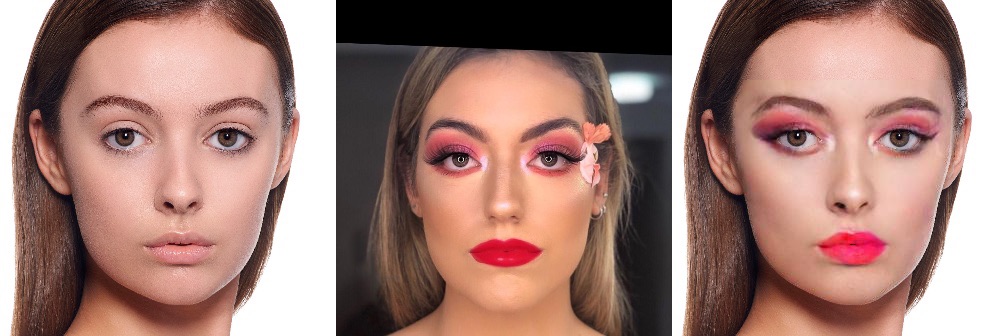}\\

\end{tabular}

\end{center}
\caption{BagNet Output Examples }
\label{fig:app:face}
\end{figure*}

\begin{figure*}[t]
\begin{center}
\begin{tabular}{c}

Input~~~~~~~~~~~~~~~~~~~~~~~~~~~~Reference~~~~~~~~~~~~~~~~~~~~~~~~~~~~~~Output \\

\midrule
\includegraphics[width=0.8\linewidth]{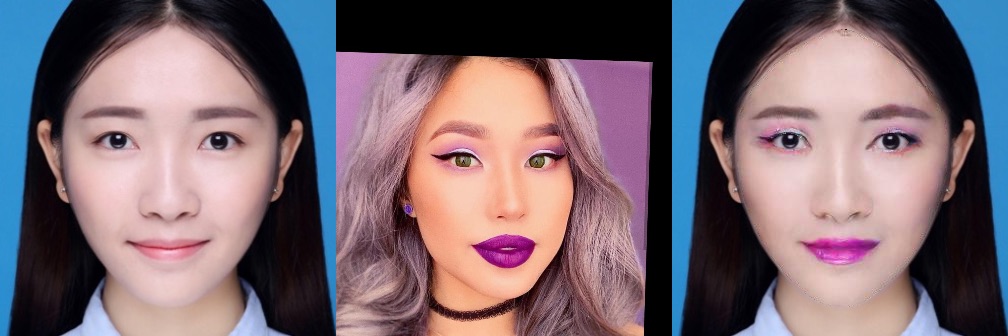}\\

\midrule
\includegraphics[width=0.8\linewidth]{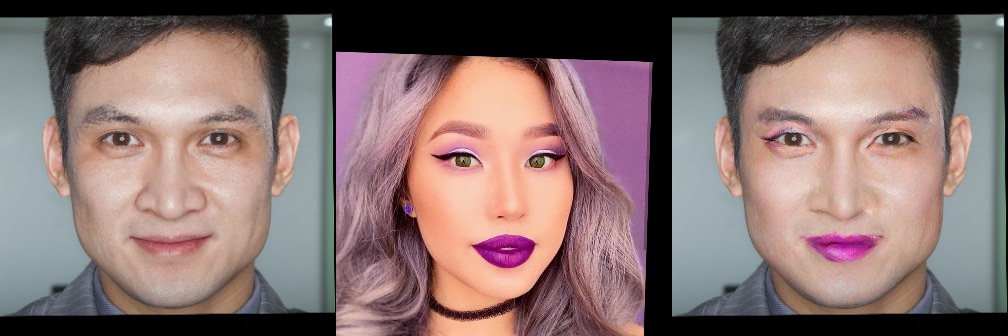}\\

\midrule
\includegraphics[width=0.8\linewidth]{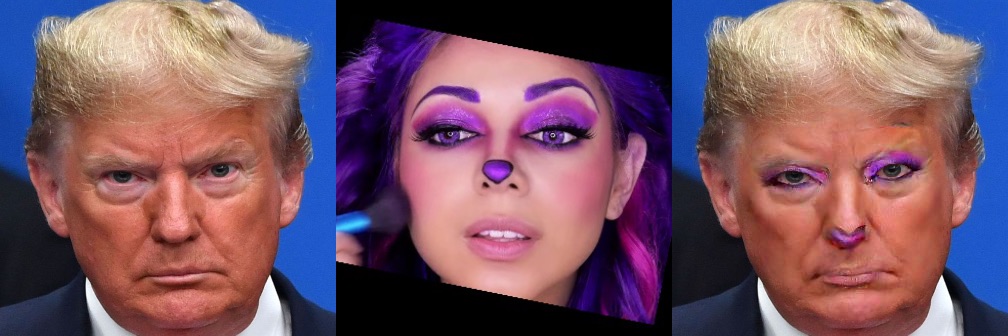}\\

\midrule
\includegraphics[width=0.8\linewidth]{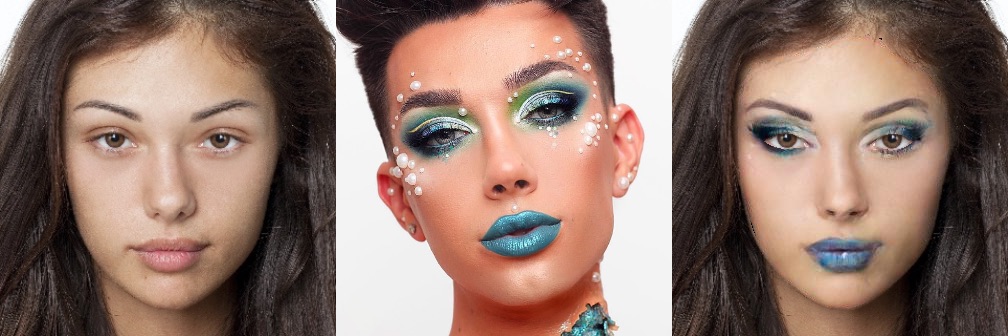}\\

\end{tabular}

\end{center}
\caption{BagNet Output Examples }
\label{fig:app:face}
\end{figure*}

\end{appendices}
\end{document}